\documentclass[runningheads]{llncs}
\usepackage{graphicx}
\usepackage{amssymb}
\usepackage{amsmath,mathtools}
\begin{document}

\renewcommand{\vec}[1]{\boldsymbol{\mathrm{#1}}}
\title{Inpainting Transformer for Anomaly Detection}
%
%
\author{Jonathan Pirnay\inst{1} \and
Keng Chai\inst{1}}
%
%
\institute{Digital Incubation, Fujitsu Technology Solutions GmbH, Munich, Germany\\
\email{\{jonathan.pirnay,keng.chai\}@fujitsu.com}}
\maketitle              
\begin{abstract}
Anomaly detection in computer vision is the task of identifying images which deviate from a set of normal images. A common approach is to train deep convolutional autoencoders to inpaint covered parts of an image and compare the output with the original image. By training on anomaly-free samples only, the model is assumed to not being able to reconstruct anomalous regions properly. For anomaly detection by inpainting we suggest it to be beneficial to incorporate information from potentially distant regions. In particular we pose anomaly detection as a patch-inpainting problem and propose to solve it with a purely self-attention based approach discarding convolutions. The proposed Inpainting Transformer (InTra) is trained to inpaint covered patches in a large sequence of image patches, thereby integrating information across large regions of the input image. When training from scratch, in comparison to other methods not using extra training data, InTra achieves results on par with the current state-of-the-art on the MVTec AD dataset for detection and surpassing them on segmentation.

\end{abstract}


\section{Introduction}
\label{section:introduction}

Anomaly detection and localization in vision describe the problem of deciding whether a given image is atypical with respect to a set of normal samples, and to identify the respective anomalous subregions within the image. Both problems have strong implications for industrial inspection \cite{MvtecAd2019} and medical applications \cite{fernando2020deep}. In practical industrial applications, anomalies occur rarely. Due to the lack of sufficient anomalous samples, and as anomalies can be of unexpected shape and texture, it is hard to deal with this problem with supervised methods. Current approaches follow unsupervised methods and try to model the distribution of normal data only. At test time an anomaly score is given to each image to indicate how much it deviates from normal samples. For anomaly localization a similar score is assigned to subregions or individual pixels of the image.

\begin{figure}[t]
	\centering
	\includegraphics[width=\columnwidth]{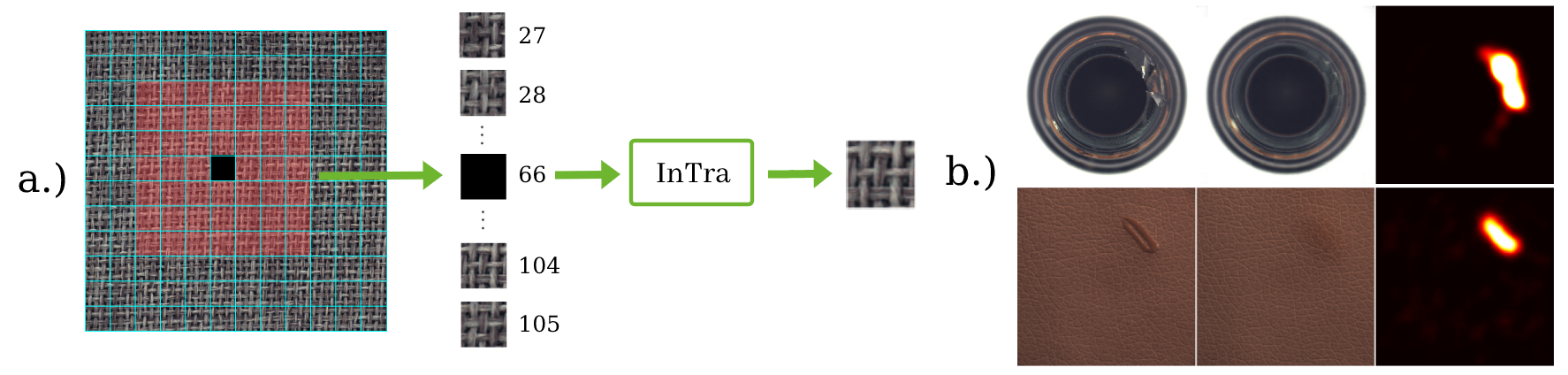}
	\caption{Schematic overview of the proposed method.
	 a.) The image is split into square patches. An inpainting transformer model (InTra) is trained to reconstruct a covered patch (black) from a long sequence of surrounding patches (red). Positional embeddings are added to the patches to include spatial context. b.) Examples: By reconstruction of all patches of an input image (left), a full reconstruction is obtained (middle). Comparison of original and reconstruction yields a pixel-wise anomaly score (right).
	}
	\label{fig:graphical_abstract}
\end{figure}

A common approach following this paradigm is to use deep convolutional autoencoders or generative models such adversarial networks in order to model the manifold of normal training data. The difference between the input and reconstructed image is then used to compute the anomaly scores. In practice this approach often suffers from the drawback that convolutional autoencoders generalize strongly and anomalies are reconstructed well, leading to misdetection. Recent methods propose to mitigate this effect by posing the generative part as an inpainting problem: Parts of the input image are covered and the model is trained to reconstruct the covered parts in a self-supervised way \cite{Bhattad2018DetectingAF,8614226,ZAVRTANIK2021107706,DBLP:journals/corr/abs-2010-01942}. By conditioning on the neighborhood of the excluded part only, small anomalies get effectively retouched. Due to their limited receptive field, fully convolutional neural networks (CNNs) are partially ineffective in modeling distant contextual information, which makes the removal of larger anomalous regions difficult. For inpainting in general settings, this can be effectively addressed by introducing contextual attention in the model \cite{Yu_2018_CVPR}. For inpainting in the context of anomaly detection we suggest it to be beneficial to learn the relevant patterns alone by combining information from large regions around the covered image part via attention.

Inspired by the recent success of self-attention based models such as Transformers \cite{NIPS2017_3f5ee243} in image recognition \cite{dosovitskiy2020}, we pose anomaly detection as a patch-inpainting problem and propose to solve it without convolutions: images are split into square patches, and a Transformer model is trained to reconstruct covered patches on the basis of a long sequence of neighboring patches. By recovering the whole image in this way, a full reconstructed image is obtained where the reconstruction of an individual patch incorporates a large context and not only the appearance of its immediate neighborhood. Thus patches are not reconstructed by simply mimicking the local neighborhood, leading to high anomaly scores even for spacious anomalous regions.

Our contributions enfold the modeling of anomaly detection as a patch-sequence inpainting problem which we solve using a deep Transformer network consisting of a simple stack of multiheaded self-attention blocks. Within this network convolutional operations are removed entirely. Furthermore we propose to employ long residual connections between the Transformer blocks and to perform a nonlinear dimension reduction for keys and queries when computing self-attention
in order to improve the network's reconstruction capabilities for difficult surfaces. By adding embeddings of the position of individual patches within an image to the sequence of patches, it is possible to perform the inpainting in a global context even if the sequence of patches does not cover the full image.

We evaluate our method on the challenging MVTec AD dataset \cite{MvtecAd2019} for both detection and segmentation. Although Transformer networks are usually trained on huge amounts of data, we effectively train our networks with $\sim$55M parameters from scratch only on the 60-400 images available for each category in MVTec AD. We compare our results to the current state-of-the-art not using any extra training data. Our proposed method InTra achieves on par results on the detection task and slightly better results on the segmentation task.

\section{Related Work}
\label{sec:rel_work}

In anomaly detection, reconstruction-based methods try to model only normal, defect-free samples. For this, deep CNN autoencoders are widely used to learn the manifold of defect-free images in a latent bottleneck. Given defective test data, these models should not be able to properly reconstruct the anomalous image since they only model normal data \cite{CHOW2020101105,Sakurada2014autencoder,Baur_2019}. An anomaly map for segmentation is usually generated via pixel-wise difference or similarity measures between the input image and its model reconstruction, leading to noticeable anomalies.

Even though in reconstruction-based methods the models are trained on defect-free samples only, they often generalize well to anomalies in practice \cite{gong2019memorizing}.
An inpainting scheme can be used to effectively hide anomalous regions to further restrict a model's capability to reconstruct anomalies \cite{Bhattad2018DetectingAF,8614226,ZAVRTANIK2021107706,DBLP:journals/corr/abs-2010-01942}. By covering parts of the original image, the reconstruction method needs to have semantic understanding of the image to be able to generate a coherent and realistic image. Zavrtanik et al. propose to use a U-Net architecture \cite{10.1007/978-3-319-24574-4_28} taking advantage of long residual connections. Their reconstruction-based method randomly selects multiple parts of the image to inpaint, yielding the current state-of-the-art results for anomaly detection via inpainting for different benchmarks \cite{ZAVRTANIK2021107706}.

Anomalies which span over a large area may still cause problems as these will not be covered up sufficiently enough.
As such we propose to add global context by replacing CNNs with a Transformer-based framework applied in vision.


Transformer models were originally introduced in natural language processing (NLP) and have since evolved to be the modern design for various sequence tasks like text translation, generation and document classification \cite{NIPS2017_3f5ee243,devlin2019bert,yang2020xlnet}.

In a Transformer model, self-attention is used to relate elements of a sequence to each other. Based on the relative weighted importance a shared representation is calculated taking into account the relative dependencies between sequence elements.
This is able to replace recurrent neural networks in sequence-to-sequence modeling because long-range dependencies are processed globally.
The general architecture can be found in the original work \cite{NIPS2017_3f5ee243}.

While Transformer architectures have been widely studied in NLP and sequence modeling, convolutional architectures have been essentially the standard tool in recent years due to weight sharing, translation equivariance and locality.
Due to the induced bias in fully convolutional autoencoders, the restricted receptive field limits global context \cite{Yu_2018_CVPR}.
Even though in theory the self-attention framework may mitigate this problem, running self-attention on the whole image without further simplifications is not feasible \cite{ho2019axial,parmar2018image}.

Recently Dosovitskiy et al. have proposed Vision Transformer \cite{dosovitskiy2020}, where the image data is split up into square non-overlapping uniform patches.
Each patch and position gets embedded into a latent space and every image is treated as a sequence of these embedded patches.
A Transformer architecture is applied on the restructured data achieving comparable results to state of the art CNNs and even surpassing them on some tasks while reducing model bias.

\section{Inpainting Transformer for Anomaly Detection}

Our approach is based on a simple stack of Transformer blocks which are trained to inpaint covered image patches based on neighboring patches. An overview of the method is shown in Figure \ref{fig:graphical_abstract}.

\subsection{Patch Embeddings and Multihead Feature Self-attention}
\label{seq:embedding_patches}

We use a similar notation as in \cite{dosovitskiy2020}. Let $\vec x \in \mathbb R^{H \times W \times C}$ be an input image, where $(H,W)$ denotes the (height, width)-size and $C$ the number of channels of the image. Let $K$ be the desired side length of a square patch and $N := \frac{H}{K}, M := \frac{W}{K}$ (the image is resized such that $K$ divides $H$ and $W$). We split the image $\vec x$ into a $N \times M$ grid of flattened square patches
\begin{equation*}
\displaystyle \vec x_p \in \mathbb R^{\displaystyle (N \times M) \times (K^2 \cdot C)},
\end{equation*}
where $\vec x^{(i,j)}_p \in \mathbb R^{K^2 \cdot C}$ is the patch in the $i$-th row and $j$-th column. Our aim is to choose square subgrids of some side length $L$ in this patch grid and train a network to reconstruct any covered patch in the subgrid based on the rest of the subgrid's patches. Formally, this inpainting problem is as follows:

Let
$
\left( \vec x_p^{(i,j)} \right)_{\displaystyle (i,j) \in S}
$
be such a square subgrid ("window") of patches defined by some index set $S = \{r, \dots, r + L-1\} \times \{s, \dots, s + L - 1\}$. Here $L$ is the side length of the window, and $(r,s)$ is the grid position of the window's upper left patch. If $(t, u) \in S$ is the position of some patch, the formal task to inpaint $(t,u)$ given $S$ is to approximate the patch $\vec x_p^{(t,u)}$ using only the content and positions of all other patches $\left( \vec x_p^{(i,j)} \right)_{(i,j) \in S \setminus \{(t,u)\}}$ in the window.

As by definition Transformers are invariant with respect to reorderings of the input, the one-dimensional positional information $f(i,j) := (i - 1) \cdot N + j$ of a patch $\vec x_p^{(i,j)}$ is used.
To use as a sequence input to the Transformer model, we map the window of patches and their positional information into some latent space of dimension $D$, i.e. for each patch $\vec x_p^{(i,j)}$ with $(i,j) \in S \setminus \{(t,u)\}$ we set

\begin{equation}
	\label{eq:patch_embedding}
	\vec y^{(i,j)} := \vec x_p^{(i,j)} \vec E + \text{posemb}(f(i,j)) \in \mathbb R^D
\end{equation}
with learnable weight matrix $\vec E \in \mathbb R^{(K^2 \cdot C) \times D}$, and where $\text{posemb}$ denotes a standard learnable one-dimensional position embeddings.

To account for the patch at position $(t,u)$ to inpaint, we add a {\em single} learnable embedding $\vec x_{\text{inpaint}} \in \mathbb R^D$ to the position embedding via
\begin{equation}
	\label{eq:inpainting_embedding}
	\vec z := \vec x_{\text{inpaint}} + \text{posemb}(f(t,u)) \in \mathbb R^D.
\end{equation}
The embedding $\vec x_{\text{inpaint}}$ is comparable to the class token in \cite{devlin2019bert}.

The vectors $\vec z$ and $\vec y^{(i,j)}$ for all $(i,j) \in S \setminus \{(t,u)\}$ build the final sequence of embedded patches which serves as an input sequence of length $L^2$ to the Inpainting Transformer model.

Applying multihead self-attention (MSA) to an input sequence forms the heart of a standard Transformer block as in \cite{NIPS2017_3f5ee243}. For this, queries $\vec q$, keys $\vec k$ and values $\vec v$ are obtained by mapping the input sequence with learnable weight matrices $\vec W_q, \vec W_k, \vec W_v \in \mathbb R^{D \times D}$. Self-attention is then computed over slices of $\vec q, \vec k, \vec v$. In cases where the patches of the training images are very similar but indistinct the dot product of queries and keys are very close to each other, leading to an almost uniform softmax-weighted sum in the calculation of MSA. To mitigate this, we propose to perform a nonlinear dimension reduction when computing $\vec q$ and $\vec k$. For this, $\vec W_q$ and $\vec W_k$ are swapped with multilayer perceptrons (MLP) with a single hidden layer. In all our models we used an output dimension of $\frac{D}{2}$ and a hidden layer dimension of $2 \cdot D$ with GELU non-linearity. We refer to this modified MSA as multihead feature self-attention (MFSA). We experienced improved detection results with MFSA (see Section \ref{sec:ablation_studies}). However, depending on the output and hidden dimension of the MLPs, the number of learnable parameters increases strongly with MFSA.

\subsection{Network Architecture and Training}

\begin{figure*}[t]
	\centering
	\includegraphics[width=0.85\columnwidth]{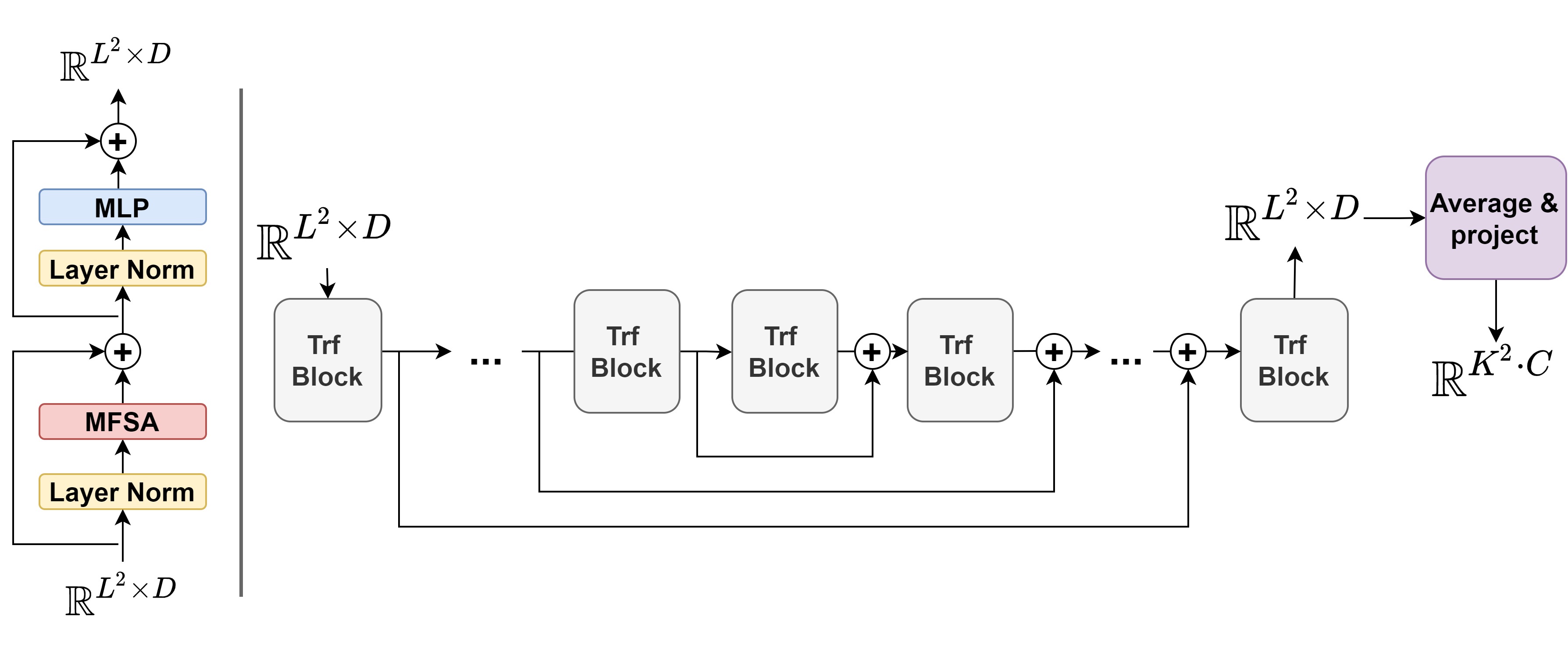}
	\caption{Overview of the proposed architecture. Left: Parts of an individual Transformer block. Right: A stack of Transformer blocks builds the full architecture. Long residual connections are used to add information from earlier blocks to later ones.}
	\label{fig:network_architecture}
\end{figure*}

Our network architecture for inpainting is composed of a simple stack of $n$ Transformer blocks. Figure \ref{fig:network_architecture} illustrates the architecture. The structure of each Transformer block mainly follows \cite{dosovitskiy2020} and consists of MFSA followed by a multilayer perceptron (MLP). Layer normalization is applied before ("pre-norm" \cite{Wang2019LearningDT}), and residual connections after MFSA and MLP. Each MLP has a single hidden layer with GELU nonlinearity and maps $\mathbb R^D \to \mathbb R^{4 \cdot D} \to \mathbb R^D$. In particular the input and output of each Transformer block is a sequence in $\mathbb R^{L^2 \times D}$ (see Fig. \ref{fig:network_architecture}).

To obtain the inpainted patch, we average over the output sequence of the last Transformer block to get a single vector in $\mathbb R^D$ which is mapped back to the pixel space of the flattened patches $\mathbb R^{K^2 \cdot C}$ via a learnable affine transformation followed by a sigmoidal.

In early experiments an inspection of the attention weights showed that a large spatial context is present in earlier layers. In addition to that, Attention Rollout \cite{attentionRollout} has been used in \cite{dosovitskiy2020} to illustrate that information across the entire input image is integrated already in the lowest layers. In order to carry this early information to deeper blocks of the network, we put additional long residual connections between early and late layers in a U-Net fashion \cite{10.1007/978-3-319-24574-4_28}. We found that the use of long residual connections leads to more structural detail in the overall reconstruction, slightly improving both detection and segmentation (see Section \ref{sec:ablation_studies}).

The network is trained by randomly sampling batches of patch windows with a fixed side length $L$ from normal image data. In each window a random patch position $(t,u)$ is chosen, which is inpainted by the network as described in the previous sections.

For the loss function, we compare the original and reconstructed patch with pixel-wise $L_2$ loss. To account for perceptual differences, we also include structural similarity \cite{Wang2004ssim} and gradient magnitude similarity \cite{10.1109/TIP.2013.2293423}.

Given an original and reconstructed patch $\vec x_p, \vec{\hat x}_p \in \mathbb R^{K \times K \times C}$, the full loss function $\mathcal L$ is given by
\begin{equation}
	\label{eq:loss}
\begin{split}
	\mathcal L(\vec x_p, \vec{\hat x}_p) = L_2(\vec x_p, \vec{\hat x}_p) & + \frac{\alpha}{K^2}\sum_{(i,j) \in K \times K}(\vec 1 - \text{GMS}_{\text{avg}}(\vec x_p, \vec{\hat x}_p)^{(i,j)}) \\
	& + \frac{\beta}{K^2}\sum_{(i,j) \in K \times K}(\vec 1 - \text{SSIM}_{\text{avg}}(\vec x_p, \vec{\hat x}_p)^{(i,j)})
\end{split}
\end{equation}
where $\alpha, \beta$ are individual scaling parameters, $\vec 1$ is a matrix of ones and $\text{GMS}_{\text{avg}}$ (resp. $\text{SSIM}_{\text{avg}}$) denotes the gradient magnitude similarity maps (resp. structural similarity maps) averaged over the color channels.

\subsection{Inference and Anomaly Detection}

The inferencing process is divided into two steps: First a complete inpainted image is generated, afterwards the difference between the reconstruction and original is used to compute a pixel-wise anomaly map.

Let $\vec x \in \mathbb R^{H \times W \times C}$ be an input image with an $N \times M$ patch grid as introduced above. For each patch position $(t,u) \in N \times M$, we choose an appropriate patch window of side length $L$ which is used as a basis to inpaint the patch at position $\vec x_p^{(t,u)}$. In particular we define the window by its upper left patch $\vec x_p^{(r,s)}$ via
\begin{align}
	r = & g(t) - \max(0, g(t) + L - N - 1), \\
	s = & g(u) - \max(0, g(u) + L - M - 1),
\end{align}
where the map $g$ is given by $g(c) := \max \left(1, c - \left\lfloor \frac{L}{2} \right\rfloor \right)$.

The above equations choose $(r,s)$ such that $(t,u)$ is as much centered in the $L \times L$ patch-window as possible. Using this window, the patch $\vec x_p^{(t,u)}$ is reconstructed by the network as described. By reconstructing all patches in the $N \times M$ grid, we obtain a full reconstruction $\vec{\hat x}$ of the whole image.

For the generation of an expressive anomaly map from $\vec x$ and $\vec{\hat x}$ we use a simplified variant of the GMS-based scheme proposed in \cite{ZAVRTANIK2021107706}. Denote by $\vec x_l$ an image $\vec x$ resized to scale $l$. Now for original and reconstructed images $\vec x, \vec{\hat x} \in \mathbb R^{H \times W \times C}$ and scale $l \in \{\frac{1}{2}, \frac{1}{4}\}$, we set
\begin{equation}
	\label{eq:bluravg}
	m_l(\vec x, \vec{\hat x}) := \text{bluravg}_l(\vec 1 - \text{GMS}_{\text{avg}}(\vec x, \vec{\hat x})) \in \mathbb R^{l \cdot H \times l \cdot W}
\end{equation}
for a scaled and smoothed version of the gradient difference. To ease notation, we denote by $\text{bluravg}_l$ the application of an averaging filter followed by some Gaussian blur operation, both with a predefined kernel size and variance. As in \cite{ZAVRTANIK2021107706}, smoothing improves robustness with respect to small, poorly reconstructed anomalous regions. We resize the two-dimensional maps $m_{\frac{1}{2}}$ and $m_{\frac{1}{4}}$ back to the image's original size and take the pixel-wise mean which yields a difference map $\text{diff}(\vec x, \vec{\hat x}) \in \mathbb R^{H \times W}$.

To finally obtain an anomaly map for $\vec x$ during inference, we take the squared deviation of the difference map to the normal training data, i.e.
\begin{equation}
	\label{eq:anomap}
	\text{anomap}(\vec x) := \left(\text{diff}(\vec x, \vec{\hat x}) - \frac{1}{\lvert T \rvert}\sum_{\vec z \in T}{\text{diff}(\vec z, \vec{\hat z})}\right)^2 \in \mathbb R^{H \times W}_{\geq 0},
\end{equation}
where $T$ is the set of normal training samples. The pixel-wise maximum of $\text{anomap}(\vec x)$ is taken as a scalar anomaly score for detection on the image level.
An example of an anomaly map can be seen in Fig. \ref{fig:graphical_abstract}b.).

\section{Experiments}
We evaluate our method on the MVTec AD dataset which contains high resolution samples of 5 texture and 10 object categories stemming from manufacturing \cite{MvtecAd2019}. The dataset has been a widely used benchmark for anomaly detection and localization in the manufacturing domain. Each category consists of around 60 to 400 normal, defect-free samples for training. For each test image there is a ground-truth binary image labeled on pixel-level for segmentation of anomalous test images.

Based on an image's anomaly score we report standard ROC AUC as a detection metric. For localisation, the image's anomaly map (\ref{eq:anomap}) is used for an evaluation of pixel-wise ROC AUC.

\begin{table}[t]
	\caption{Detection/Segmentation results for MVTec AD. Results are presented in ROC AUC $\%$ on image level for detection, and on pixel level for segmentation.}
	\label{table:results}
	\begin{center}
		\begin{tabular}{l c c c c c}
			\hline
			Category & RIAD \cite{ZAVRTANIK2021107706} & CutPaste \cite{li2021cutpaste} & InTra (Ours) & InTra Image Size \\
			\hline \hline
			& Det. / Seg. & Det. / Seg. & Det. / Seg. & \\
			Carpet & 84.2 / 96.3 & 93.1 / 98.3 & \textbf{98.8} / \textbf{99.2} & 512 $\times$ 512 \\
			Grid & 99.6 / \textbf{98.8} & 99.9 / 97.5 &  \textbf{100.0} / \textbf{98.8} & 256 $\times$ 256 \\
			Leather & \textbf{100.0} / 99.4 & \textbf{100.0} / \textbf{99.5} & \textbf{100.0} / \textbf{99.5} & 512 $\times$ 512  \\
			Tile & 93.4 / 89.1 & 93.4 / 90.5 & \textbf{98.2} / \textbf{94.4} & 512 $\times$ 512 \\
			Wood & 93.0 / 85.8 & \textbf{98.6} / \textbf{95.5} & 97.5 / 88.7 & 512 $\times$ 512 \\
			\hline
			avg. textures & 95.1 / 93.9 & 97.0 / \textbf{96.3} & \textbf{98.9} / 96.1 \\
			\hline
			Bottle & 99.9 / \textbf{98.4} & 98.3 / 97.6 & \textbf{100.0} / 97.1 & 256 $\times$ 256 \\
			Cable & \textbf{81.9} / \textbf{94.2} & 80.6 / 90.0 & 70.3 / 91.0 & 256 $\times$ 256 \\
			Capsule & 88.4 / 92.8 & \textbf{96.2} / 97.4 & 86.5 / \textbf{97.7} & 320 $\times$ 320 \\
			Hazelnut & 83.3 / 96.1 & \textbf{97.3} / 97.3 & 95.7 / \textbf{98.3} & 256 $\times$ 256 \\
			Metal Nut & 88.5 / 92.5 & \textbf{99.3} / 93.1 & 96.9 / \textbf{93.3} & 256 $\times$ 256 \\
			Pill & 83.8 / 95.7 & \textbf{92.4} / 95.7 & 90.2 / \textbf{98.3} & 512 $\times$ 512 \\
			Screw & 84.5 / 98.8 & 86.3 / 96.7 & \textbf{95.7} / \textbf{99.5} & 320 $\times$ 320  \\
			Toothbrush & \textbf{100.0} / \textbf{98.9} & 98.3 / 98.1 & \textbf{100.0} / \textbf{98.9} & 256 $\times$ 256 \\
			Transistor & 90.9 / 87.7 & 95.5 / 93.0 & \textbf{95.8} / \textbf{96.1} & 256 $\times$ 256 \\
			Zipper & 98.1 / 97.8 & 99.4 / \textbf{99.3} & \textbf{99.4} / 99.2 & 512 $\times$ 512 \\
			\hline
			avg. objects & 89.9 / 94.3 & \textbf{94.3} / 95.8 & 93.0 / \textbf{96.9} \\
			\hline
			avg. all categories & 91.7 / 94.2 & \textbf{95.2} / 96.0 & 95.0 / \textbf{96.6} \\
			\hline
		\end{tabular}
	\end{center}
\end{table}

\subsection{Implementation Details}
\label{seq:implementation_details}

We train our model on each product category from scratch. We randomly choose 10\% of images from the normal training data (however a maximum of 20) and use them as a validation set to control the quality of reconstructions. In each epoch 600 patch windows are sampled randomly per image. To augment the dataset, random rotation and flipping is used.

The choice of three parameters has an obvious significant impact on the performance: Side length $K$ of square patches, side length $L$ of a patch window and the choice of height $H$ and width $W$ (with $H = W$, as all images are square) to which the original image is resized during training and inference. The patch size determines how much of the image is covered, the size of the patch window determines the dilation of context we include during inpainting, the image size implicitly influences both. For all models we choose $K = 16$, $L = 7$. For the choice of image size in our pipeline, a balance needs to be struck between enlarging the image context of the $7 \times 7$ window, quality of patch reconstructions and computation time, as Transformer models usually take a long time to train. The heuristics is to choose the image as small as possible while keeping patch reconstructions at a high level of detail. Hence we train the model with image dimensions $256 \times 256, 320 \times 320, 512 \times 512$ for 200 epochs and compare the best (epoch-wise) validation losses (averaged over $\pm 5$ epochs). If there is no significant improvement in the validation loss of at least $10^{-4}$ for an image dimension with the next in size, the smaller dimension is chosen. The rightmost column of Table \ref{table:results} shows the resulting image sizes for each category. We note that in practice $K, L, H, W$ could be tuned for the detection task at hand if prior knowledge about possible defects is present.

The Inpainting Transformer model trained consists of 13 blocks with 8 attention heads each and a latent dimension of $D = 512$, using MFSA. In total this amounts to $\sim$55M learnable parameters.

Given an image size of $512 \times 512$, a kernel size of $21$ (resp. $11$) is used for averaging and Gaussian blur (with $\sigma = 2$) for $\text{bluravg}_{\frac{1}{2}}$ (resp. $\text{bluravg}_{\frac{1}{4}}$) in (\ref{eq:bluravg}). The kernel sizes are scaled linearly for smaller image sizes. Anomaly maps are resized back to their original high image resolution for proper segmentation comparison.

For the loss function $\mathcal L$ in (\ref{eq:loss}) we set $\alpha = \beta = 0.01$. The network is trained using the Adam optimizer with a learning rate of $0.0001$ and a batch size of 256 until no improvement on the validation loss is observed for 50 consecutive epochs. The model weights at the epoch with the best validation loss are chosen for evaluation.
Although common for training Transformers, we don't apply dropout at any point.

\subsection{Results and Discussion}

The results for detection and segmentation are reported in Table \ref{table:results}. We compare our method to RIAD \cite{ZAVRTANIK2021107706}, which also uses an inpainting reconstruction-based method and computes anomaly maps based on GMS. Furthermore we compare the results to CutPaste \cite{li2021cutpaste} which uses a special data augmentation strategy to train a one-class classifier in a self-supervised way. CutPaste also offers results using pretrained representations, however we focus on the results without extra training data in accordance to our training procedure.

To our knowledge RIAD offers the current best performing model based on an inpainting scheme, whereas CutPaste is the best performing model on the MVTec AD benchmark not using extra training data. Our method outperforms RIAD on both detection and segmentation. On the detection task, CutPaste is superior to our method by 0.2\%, however on the segmentation task, we can improve the result by 0.6\%.

It is worth noting that there are two strongly underperforming categories: \textit{Cable} contains many anomalous images where the defect lies in the overall constitution of the product (such as missing pieces). Combined with noise in large areas this makes these anomalies hard to detect via inpainting. Although the defects in \textit{capsule} are per-se easily visible on the generated anomaly maps, our method does not learn the reconstruction of the typography sufficiently well, leading to high anomaly scores also on normal samples.

\subsection{Ablation Studies}
\label{sec:ablation_studies}

We examine the influence of certain building blocks in the architecture. All categories except \textit{Leather} are trained for 200 epochs with settings as described in Section \ref{seq:implementation_details}, if not stated otherwise. \textit{Leather} takes the longest until details start to show in the inpainted patches, so for comparability the network is trained for 700 epochs. It should be noted that training for only 200 epochs for most categories does not lead to results comparable to Table \ref{table:results}. The average results of the ablation studies are reported in Table \ref{table:ablation}. We observe a decline in both detection and segmentation when omitting long residual connections. We furthermore examine the effect of using  normal multihead self-attention (MSA) instead of MFSA as described in Section \ref{seq:embedding_patches}. Segmentation results do not improve using MFSA over MSA, however the average detection results improve by 0.7\% when using MFSA. Lastly we test how the side length $L$ of a patch window influences the performance by training the network also with $L \in \{5, 9\}$. Detection and segmentation improve with growing patch windows, as more information from distant pixels can be used for the inpainting task. This comes with high computational cost however, as the computation of the dot product in self-attention is quadratic in the sequence length.

\begin{table}[t]
	\caption{Detection/Segmentation results for the ablation studies. 'Regular' refers to the architecture as described in the previous sections, and 'NLR' to 'no long residual connections'. }
	\label{table:ablation}
	\begin{center}
		\begin{tabular}{l | c c c c c}
			\hline
			& Regular & \ \ \ \ \ \ \ NLR \ \ \ \ \ \ & MSA & \ \ \ \ \ \ \ $L = 5$ \ \ \ \ \ \ \  & $L = 9$ \\
			\hline \hline
			& Det. / Seg. & Det. / Seg. & Det. / Seg. & Det. / Seg. & Det. / Seg. \\
			avg. text. & 98.2 / 95.3 & 98.6 / 95 & 98.4 / 95.5 & 98.4 / 95.8 & 98.3 / 96.2 \\
			\hline
			avg. obj. & 90.7 / 95.5 & 90.0 / 95.3 & 89.6 / 95.5 & 90.0 / 95.2 & 90.8 / 95.8 \\
			\hline
			avg. all & 93.2 / 95.4 & 92.9 / 95.2 & 92.5 / 95.5 & 92.8 / 95.4 & 93.3 / 95.9 \\
			\hline
			diff. to Reg. & & \textbf{-0.3 / -0.2} & \textbf{-0.7 / +0.1} & \textbf{-0.5 / 0.0} & \textbf{+0.1 / +0.5} \\
		\end{tabular}
	\end{center}
\end{table}

\section{Conclusion}
\begin{samepage}
Inspired by the success of self-attention in vision tasks, we have used a Transformer model for visual anomaly detection by using an inpainting reconstruction approach. We argued that by discarding convolutions and using only self-attention to incorporate global context into reconstructions, anomalies can be successfully detected and localized. Hyperparameters such as the input image size, patch sequence length and patch dimension have a strong impact on the overall performance, and including the detection of good values for them in the training pipeline is paramount. With a simple pipeline as proposed, we have shown that InTra can reach state-of-the-art results on the popular MVTec AD dataset not using extra training data.
\end{samepage}

\bibliographystyle{splncs04}
\bibliography{intra-iciap}

\newpage
\appendix

\section{Appendix}

\begin{figure}[ht]
	\centering
	\includegraphics[width=\columnwidth]{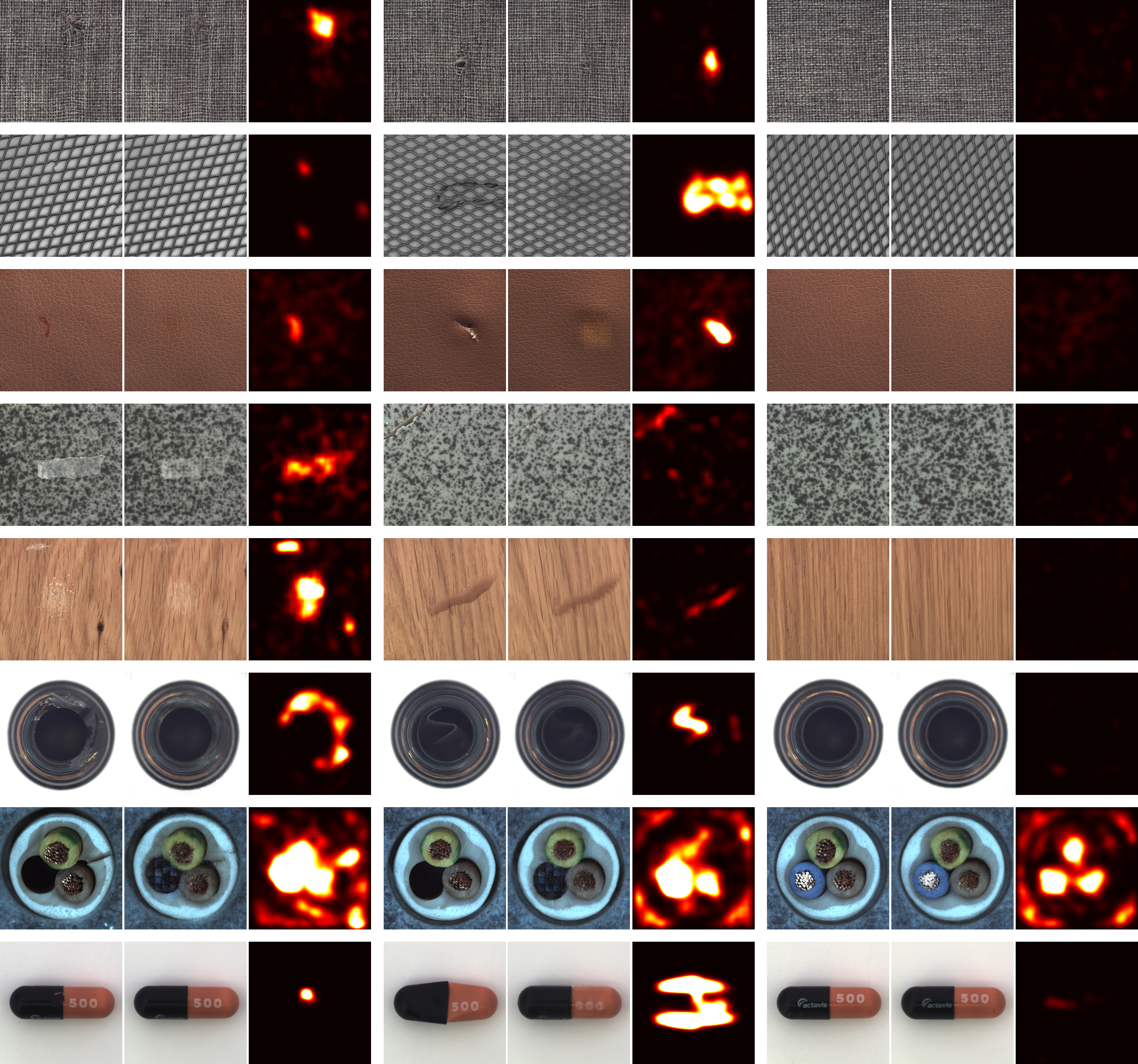}
	\caption{Qualitative results on MVTec AD for our method across different categories. Each row shows examples of one category, each column a group of three images with original (left), reconstruction (center), anomaly map (right). The two left columns show examples of anomalous test images, the rightmost column shows an example of a good test image. Categories from top to bottom: \textit{carpet, grid, leather, tile, wood, bottle, cable, capsule}. The rest of the categories are shown in Figure \ref{fig:overview_anomalies_all_pt2}.}
	\label{fig:overview_anomalies_all_pt1}
\end{figure}

\begin{figure}[ht]
	\centering
	\includegraphics[width=\columnwidth]{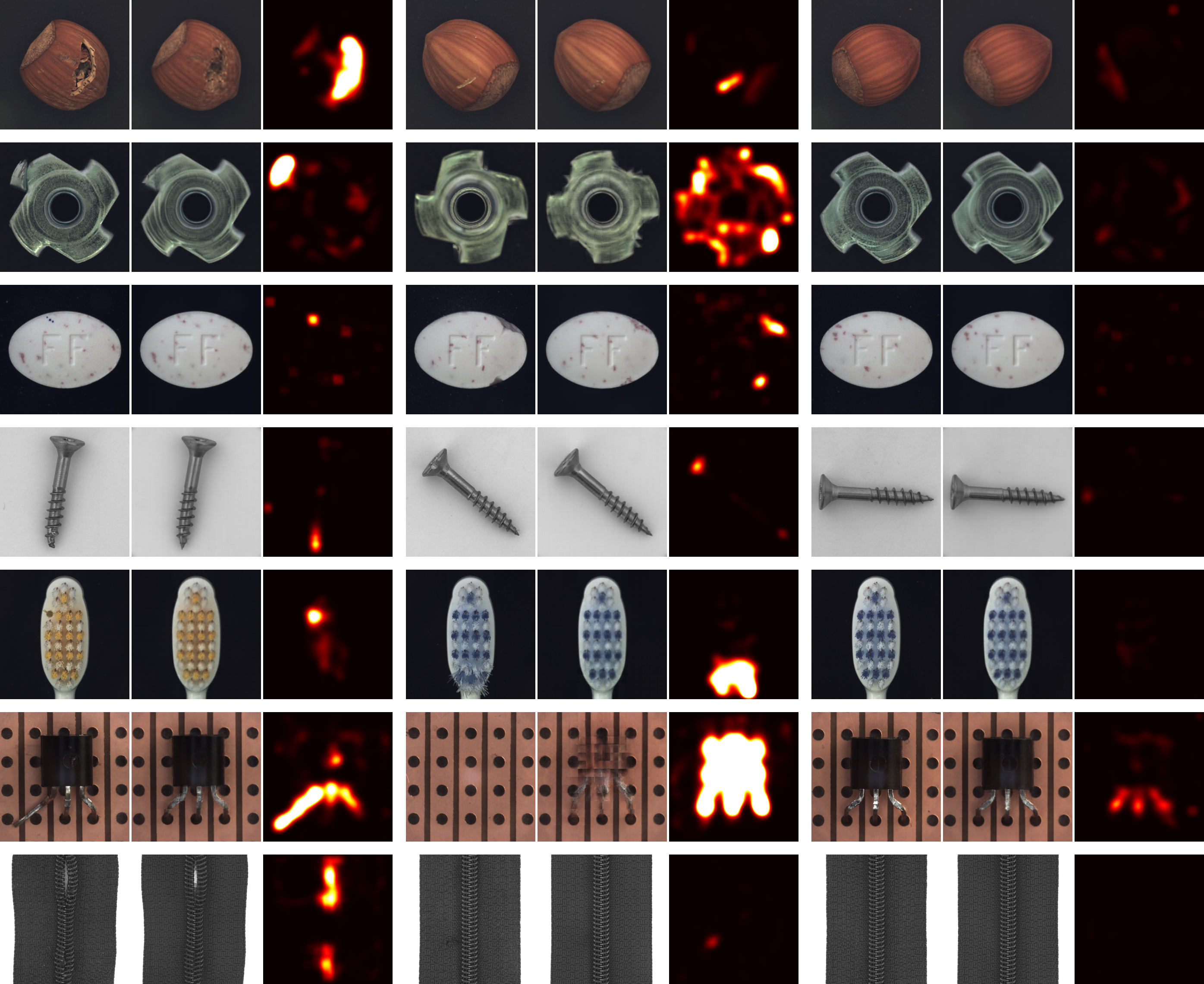}
	\caption{Examples of qualitative results continued from Figure \ref{fig:overview_anomalies_all_pt1}. Categories from top to bottom: \textit{hazelnut, metal nut, pill, screw, toothbrush, transistor, zipper}.}
	\label{fig:overview_anomalies_all_pt2}
\end{figure}





\end{document}